\documentclass[letterpaper, 10pt, journal, twoside]{IEEEtran}

\makeatletter
\let\NAT@parse\undefined
\makeatother

\usepackage{dblfloatfix}
\usepackage[numbers,sectionbib,sort]{natbib}
\usepackage{bm}
\usepackage{gensymb}
\usepackage{xcolor}
\usepackage{graphicx}
\usepackage{float}
\usepackage{amsmath}
\usepackage{amssymb}
\usepackage{subcaption}
\usepackage{amsfonts}
\usepackage{siunitx}
\usepackage{booktabs}
\usepackage{makecell}
\usepackage{multirow}
\usepackage{subfiles}
\usepackage{upgreek}
\usepackage[font=small]{caption}
\usepackage[export]{adjustbox}
\usepackage{tikz}
\usepackage{tabularx}
\usepackage{svg}
\usepackage{sidecap} \sidecaptionvpos{figure}{c}
\usepackage[hidelinks]{hyperref}
\newcommand\blfootnote[1]{%
    \begingroup
    \renewcommand\thefootnote{}\footnote{#1}%
    \addtocounter{footnote}{-1}%
    \endgroup
}

\usepackage[nameinlink, capitalize]{cleveref}
\usepackage{color}
\definecolor{CommentPink}{rgb}{1,0.2,0.5}
\definecolor{CommentBlue}{rgb}{0,0,1}
\definecolor{CommentGreen}{rgb}{0,1,0}

\Crefname{section}{Sec.}{Sec.}
\Crefname{equation}{Eq.}{Eq.}

\usepackage[printonlyused,withpage,nolist,nohyperlinks]{acronym}

\IEEEoverridecommandlockouts 

\title{ActiveGS: Active Scene Reconstruction Using Gaussian Splatting}
\author{Liren Jin$^{1}$, Xingguang Zhong$^{1}$, Yue Pan$^{1}$, Jens Behley$^{1}$, Cyrill Stachniss$^{1,3}$ and Marija Popovi\'{c}$^{2}$}

\begin{document}

\twocolumn[{
\renewcommand\twocolumn[1][]{#1}%
\maketitle
\vspace{-1.2cm}
\begin{figure}[H]
    \hsize=\textwidth
    \centering
    \includegraphics[width=0.95\textwidth]{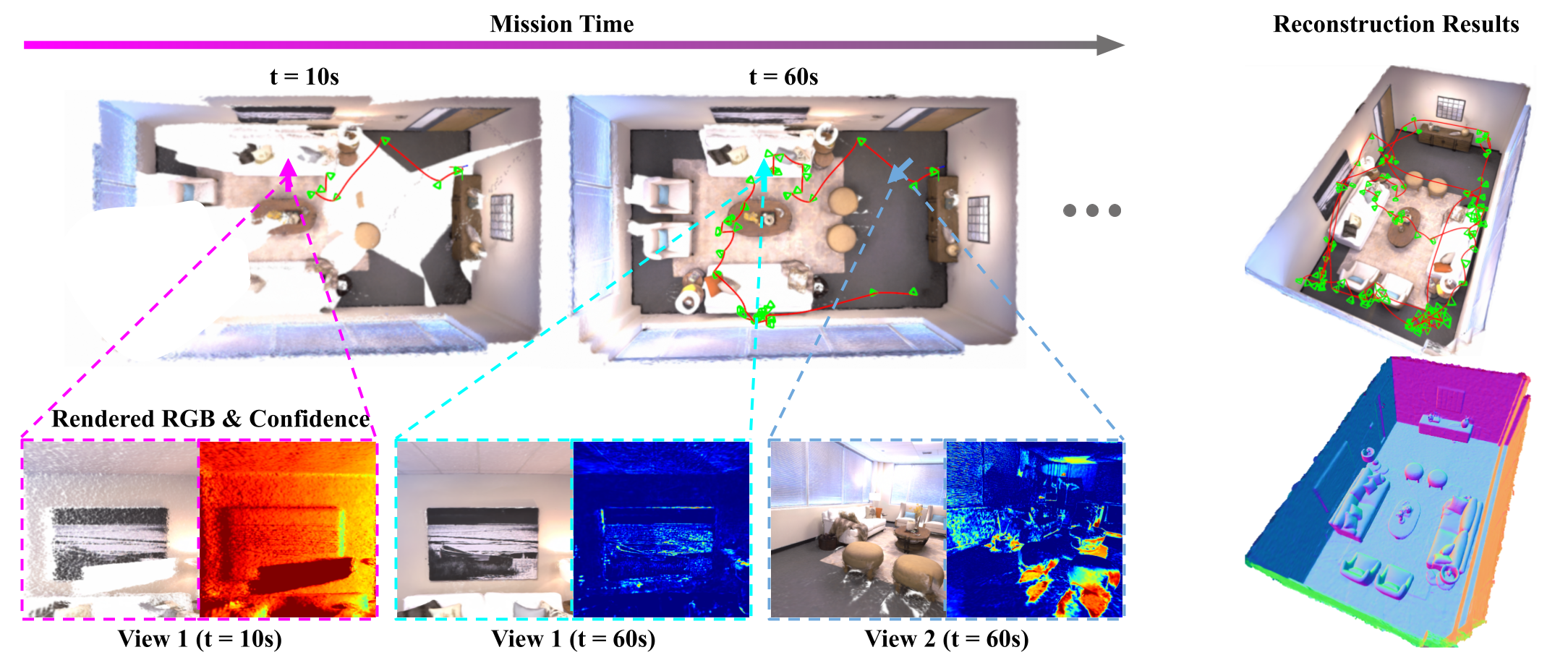}
    \caption{Our approach actively reconstructs an unknown scene. 
    We illustrate the reconstruction progress over mission time, displaying planned camera viewpoints (green pyramids) and paths (red lines). 
    We present examples of RGB and confidence maps (redder colour indicates lower confidence) rendered at the same viewpoint at different mission times (magenta and cyan arrows) and at two distinct viewpoints at the same mission time (cyan and blue arrows). 
    By integrating confidence modelling into the Gaussian splatting pipeline, our approach enables targeted view planning to build a high-fidelity Gaussian splatting map. 
    The complete camera path and final reconstruction results, including RGB rendering and surface mesh, are visualised on the right.}
    \label{F:teaser}
    \vspace{-0.4cm}
\end{figure}
}]

\markboth{IEEE Robotics and Automation Letters. Preprint Version. Accepted March, 2025}{Jin \MakeLowercase{\textit{et al.}}: ActiveGS}
\blfootnote{Manuscript received: December 20, 2024; Revised: February 23, 2025; Accepted: March 20, 2025.}
\blfootnote{This paper was recommended for publication by Editor Javier Civera upon evaluation of the Associate Editor and Reviewers’ comments. This work has been fully funded by the Deutsche Forschungsgemeinschaft (DFG, German Research Foundation) under Germany's Excellence Strategy, EXC-2070 -- 390732324 (PhenoRob).}
\blfootnote{$^{1}$L. Jin, X. Zhong, Y. Pan, J. Behley, C. Stachniss are with the Center for Robotics, University of Bonn, Germany. $^{2}$M. Popovi\'{c} is with the MAVLab, TU Delft, the Netherlands. $^{3}$C. Stachniss is also with the Lamarr Institute for Machine Learning and Artificial Intelligence, Germany.}
\blfootnote{Digital Object Identifier (DOI): see top of this page.}

\begin{abstract}
Robotics applications often rely on scene reconstructions to enable downstream tasks. 
In this work, we tackle the challenge of actively building an accurate map of an unknown scene using an RGB-D camera on a mobile platform. 
We propose a hybrid map representation that combines a Gaussian splatting map with a coarse voxel map, leveraging the strengths of both representations: the high-fidelity scene reconstruction capabilities of Gaussian splatting and the spatial modelling strengths of the voxel map.
At the core of our framework is an effective confidence modelling technique for the Gaussian splatting map to identify under-reconstructed areas, while utilising spatial information from the voxel map to target unexplored areas and assist in collision-free path planning.
By actively collecting scene information in under-reconstructed and unexplored areas for map updates, our approach achieves superior Gaussian splatting reconstruction results compared to state-of-the-art approaches. 
Additionally, we demonstrate the real-world applicability of our framework using an unmanned aerial vehicle.
\end{abstract}

\section{Introduction} \label{S:introduction}
\IEEEPARstart{A}{ctive} exploration and reconstruction of unknown scenes are relevant capabilities for developing fully autonomous robots~\citep{chen2011ijrr}. 
For scenarios, including search and rescue, agricultural robotics, and industrial inspection, online active reconstruction using mobile robots demands both mission efficiency and reconstruction quality. 
Achieving this requires two key components: high-fidelity map representations for modelling fine-grained geometric and textural details of the scenes and adaptive view planning strategies for effective sensor data acquisition. 

In this work, we tackle the problem of actively reconstructing unknown scenes using posed RGB-D camera data. 
Given a limited budget, e.g. mission time, our goal is to obtain an accurate 3D representation of the scene by actively positioning the robot with its camera online during a mission. 
Existing active scene reconstruction frameworks mainly rely on conventional map representations such as voxel grids, meshes, or point clouds~\citep{kompis2021ral,song2018icra,song2017icra,bircher2016icra,schmid2020ral,zhou2021ral,zeng2020iros}. However, these methods often do not deliver high-fidelity reconstruction results due to their sparse representations. 
Recent advances in implicit neural representations, e.g. neural radiance fields (NeRFs)~\citep{mildenhall2020eccv}, have attracted significant research interest for their accurate dense scene reconstruction capabilities and low memory footprints. 
In the context of active reconstruction, several emerging works~\citep{pan2022eccv,feng2024cvpr, jin2023iros, jin2024iros1,ran2023ral, yan2023ral} incorporate uncertainty estimation in NeRFs and exploit it to guide view planning. 
While these approaches demonstrate promising results, the rather costly volumetric rendering procedure during online incremental mapping poses limitations for NeRF-based active scene reconstruction. 

Dense reconstruction using \mbox{Gaussian splatting (GS)~\citep{kerbl2023tog}} offers a promising alternative to NeRFs, addressing rendering inefficiencies while preserving representation capabilities.
GS explicitly models scene properties through Gaussian primitives and utilises efficient differentiable rasterisation to achieve dense reconstruction. 
Its fast map updates and explicit structure make it well-suited for online incremental mapping. 
Building on these strengths, we adopt GS for active scene reconstruction in this work.

Incorporating GS into an active scene reconstruction pipeline presents significant challenges. 
First, active reconstruction often requires evaluating the reconstruction quality to guide view planning. 
However, this is difficult without ground truth information at novel viewpoints. 
Second, the Gaussian primitives represent only occupied space, making it hard to distinguish between unknown and free space, which are important for exploration and path planning.

The key contribution of this work is addressing these challenges through our novel GS-based active scene reconstruction framework, denoted as ActiveGS.
To tackle the first challenge, we propose a simple yet effective confidence modelling technique for Gaussian primitives based on viewpoint distribution, enabling view planning for inspecting under-reconstructed surfaces. 
For the second challenge, we combine the GS map with a conventional coarse voxel map, exploiting the strong representation capabilities of GS for scene reconstruction with the spatial modelling strengths of voxel maps for exploration and path planning.
 
We make the following three claims: 
(i) our ActiveGS framework achieves superior reconstruction performance compared to state-of-the-art NeRF-based approach and GS-based baselines; 
(ii) our explicit confidence modelling for Gaussian primitives enables informative viewpoint evaluation and targeted inspection around under-reconstructed surfaces, further improving mission efficiency and reconstruction quality;
(iii) we validate our approach in different synthetic indoor scenes and in a real-world scenario using an unmanned aerial vehicle (UAV). 
To support reproducibility and future research, we open-source our implementation code at: \mbox{\url{https://github.com/dmar-bonn/active-gs}}.

\section{Related Work} \label{S:related_work}
Our work uses Gaussian splatting as the map representation for active scene reconstruction. 
In this section, we overview state-of-the-art high-fidelity map representations, focusing on GS and active scene reconstruction methods.

\subsection{Gaussian Splatting as Map Representation} \label{SS:gaussian_splatting_as_map_representation}
Conventional map representations such as voxel grids~\citep{hornung2013ar}, meshes~\citep{whelan2015rss,behley2018rss}, and point clouds~\citep{zeng2020iros} often only capture coarse scene structures, struggling to provide fine-grained geometric and textural information crucial for many robotics applications~\citep{chen2011ijrr}. 
Recent advances in implicit neural representations, such as NeRFs~\citep{mildenhall2020eccv,muller2022tog}, show promising results in high-fidelity scene reconstruction by modelling scene attributes continuously. 
Although they achieve impressive reconstruction results, NeRFs require dense sampling along rays for view synthesis, a computationally intensive process that limits their online applicability.

3D GS~\citep{kerbl2023tog} offers an efficient alternative for high-fidelity scene reconstruction by combing explicit map structures with volumetric rendering. 
Unlike NeRFs, 3D GS stores scene information using explicit 3D Gaussian primitives, eliminating the need for inefficient dense sampling during volumetric rendering. 
This explicit nature also makes it well-suited for online incremental mapping, which requires frequent data fusion and scene attributes modification.
Follow-up works further enhance geometric quality by regularising 3D GS training~\citep{guedon2024cvpr} or directly adopting 2D GS for improved surface alignment~\citep{dai2024siggraph,huang2024siggraph}. 
2D GS collapses the 3D primitive volume into 2D oriented planar Gaussian primitives, enabling more accurate depth estimation and allowing the integration of normal information into the optimisation process. 
Motivated by its strong performance, we utilise 2D GS, specifically Gaussian surfel~\citep{dai2024siggraph}, as our GS map representation for active scene reconstruction.

\subsection{Active Scene Reconstruction} \label{SS:active_scene_reconstruction}
Active scene reconstruction using autonomous mobile robots is an area of active research~\citep{chen2011ijrr}. 
Given an unknown scene, the goal is to explore and map the scene by actively planning the robot's next viewpoints for effective data acquisition.
Existing active scene reconstructions utilise map representations such as voxel maps~\citep{isler2016icra, bircher2016icra,schmid2020ral,zhou2021ral,kompis2021ral}, meshes~\citep{song2018icra,song2017icra}, or point clouds~\citep{zeng2020iros, gao2022iros}. 
These approaches primarily focus on fully covering the unknown space, rather than achieving high-fidelity reconstruction.
However, high-fidelity scene reconstruction is crucial for downstream robotic tasks that rely on accurate map information.

To address this, recent research explores implicit neural representations for active reconstruction applications.
In an object-centric setup, several methods incorporate uncertainty estimation into NeRFs~\citep{pan2022eccv,jin2023iros, jin2024iros1, yan2023ral,ran2023ral} and use this information to select next best viewpoints.
For scene-level reconstruction,
\citet{yan2023iccv} and \citet{kuang2024iros} investigate the loss landscape of implicit neural representations during training to identify under-reconstructed areas.
NARUTO~\citep{feng2024cvpr} learns an uncertainty grid map alongside a hybrid neural scene representation, guiding data acquisition in uncertain regions.
These implicit neural representations often face challenges such as inefficient map updates and catastrophic forgetting during incremental mapping.

Several concurrent works propose using GS for active scene reconstruction. GS-Planner~\citep{jin2024iros} and HGS-planner~\citep{xu2024arxiv} incorporate unknown voxels into the GS rendering pipeline and detect unseen regions for exploration. 
\citet{li2024arxiv} use a Voronoi graph to extract a traversable topological map from the GS representation for path planning. 
The approach is designed for a 2D planning space, reducing its effectiveness in cluttered environments.
FisherRF~\citep{jiang2023eccv} evaluates the information content of a novel view by measuring the Fisher information value in the GS parameters. 
This procedure requires computationally expensive gradient calculations at each previously visited and candidate viewpoint, making view planning inefficient for online missions.
We build upon the idea of using GS for active scene reconstruction, while introducing a key innovation: we explicitly model the confidence of each Gaussian primitive, enabling viewpoint sampling around low-confidence Gaussian primitives for targeted inspection and fast feed-forward confidence rendering for efficient viewpoint evaluation.

\section{Our Approach} \label{S:our_approach}

\begin{figure}[!t]
  \includegraphics[width=\columnwidth]{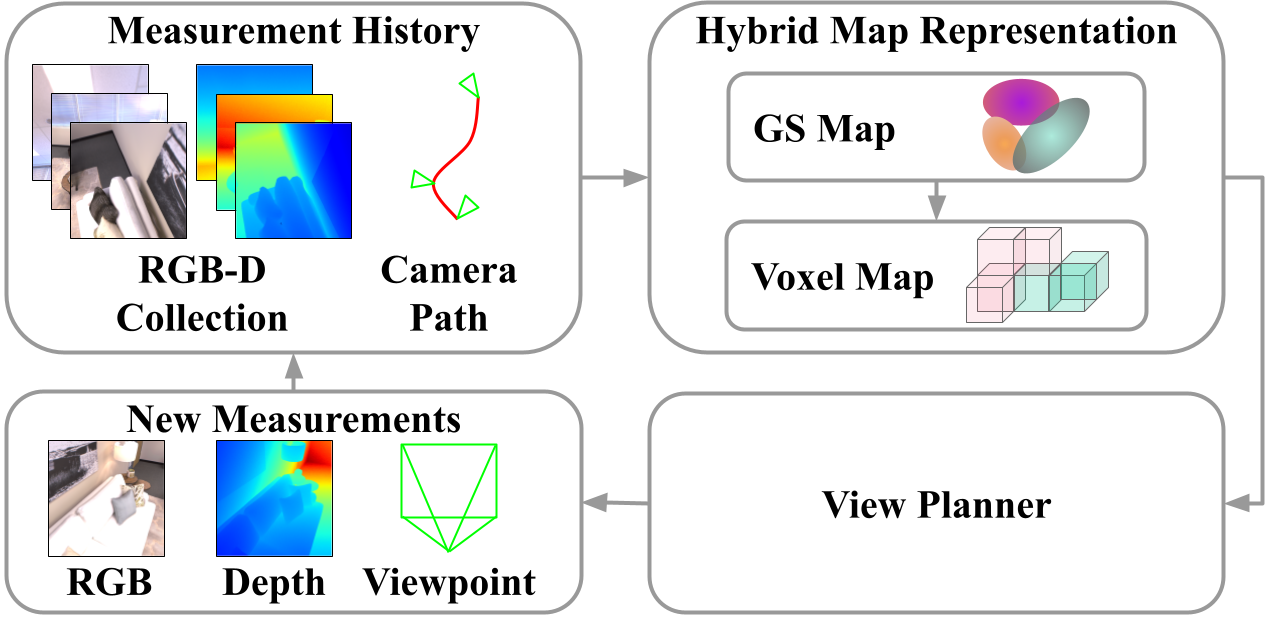}
  \caption{An overview of our framework. Our hybrid map representation consists of a 2D GS map for high-fidelity scene reconstruction with a coarse voxel map for exploration and path planning. 
  Our view planner leverages unexplored regions in the voxel map for exploration and low-confidence Gaussian primitives for targeted inspection, collecting informative measurements at planned viewpoints for map updates.
  We iterate between the map update and view planning steps until the pre-allocated mission time is reached.} \label{F:framework}
  \vspace{-0.1cm}
\end{figure}

We introduce ActiveGS, a novel framework for active scene reconstruction using GS for autonomous robotic tasks. 
An overview of our framework is shown in~\cref{F:framework}. Our goal is to reconstruct an unknown scene using a mobile robot, e.g. a UAV, equipped with an on-board RGB-D camera. 
Given posed RGB-D measurements as input, we update a coarse voxel map to model the spatial occupancy and incrementally train a GS map for high-fidelity scene reconstruction. 
To actively guide view planning to reconstruct the scene in a targeted manner, we propose using our confidence modelling technique in the GS map and information about unexplored regions in the voxel map as the basis for planning.
The framework alternates between mapping and planning steps until a predefined mission time is reached.

\subsection{Hybrid Map Representation} \label{SS:hybrid_map_representation}
Given the bounding box of the scene to be reconstructed, we uniformly divide the enclosed space into voxels, forming our voxel map $\mathcal{V}$, where each voxel $v_i \in \mathcal{V}$ represents the volume occupancy probability in the range $\left[0\,,1\right]$.

Our GS map is based on Gaussian surfel~\citep{dai2024siggraph}, a state-of-the-art 2D GS representation. 
The GS map $\mathcal{G}$ comprises a collection of Gaussian primitives.
Each primitive $\bm{g}_i \in \mathcal{G}$ is defined by its parameters $\bm{g}_i = (\bm{x}_i\,,\bm{q}_i\,,\bm{s}_i,\,\bm{c}_i\,,o_i\,,k_i)$, where $\bm{x}_i \in \mathbb{R}^{3}$ denotes the position of the primitive centre; $\bm{q}_i \in \mathbb{R}^{4}$ is its rotation in the form of a quaternion; $\bm{s}_i = \left[s^x_{i}\,,s^y_{i}\right]\in \mathbb{R}_+^{2}$ are the scaling factors along the two axes of the primitive; $\bm{c}_i \in \left[0\,,1\right]^{3}$ represents the RGB colour; $o_i \in \left[0\,,1\right]$ is the opacity; and $k_i \in \mathbb{R}_+$ is the confidence score introduced later in~\cref{SS:confidence_modelling_for_gaussian_primitives}.
The distribution of the Gaussian primitive $\bm{g}_i$ in world coordinate is represented as: 
\begin{align}
\mathcal{N}(\bm{x}; \bm{x}_i, \mathbf{\Sigma}_i) = \mathrm{exp}\left(-\frac{1}{2}(\bm{x} - \bm{x}_i)^\top \mathbf{\Sigma}_i^{-1} (\bm{x} - \bm{x}_i)\right)\,,\label{E:equation1}
\end{align}
where $\mathbf{\Sigma}_i = \mathbf{R}(\bm{q}_i) \, \mathrm{diag}\left((s^x_{i})^2, (s^y_{i})^2, 0\right) \, \mathbf{R}(\bm{q}_i)^\top$ is the covariance matrix, with $\mathbf{R}(\bm{q}_i) \in \mathit{SO}(3)$ as the rotation matrix derived from the corresponding quaternion $\bm{q}_i$ and $\mathrm{diag}(\,\cdot\,)$ indicating a diagonal matrix with the specified diagonal elements. 
The normal of the Gaussian primitive can be directly obtained from the last column of the rotation matrix as $\bm{n}_i = \mathbf{R}(\bm{q}_i)_{:, 3}$. 

Given the GS map, we can render the colour map $\mathbf{I}$, depth map $\mathbf{D}$, normal map $\mathbf{N}$, opacity map $\mathbf{O}$, and confidence map $\mathbf{K}$ at a viewpoint using the differentiable rasterisation pipeline~\citep{dai2024siggraph}. 
Without loss of generality, the rendering function for a pixel $\bm{u}$ on the view is formulated as:
\begin{align}
        \mathbf{O}(\bm{u}) &= \sum_{i=1}^{n} w_i\,, \mathbf{M}(\bm{u}) = \sum_{i=1}^{n} w_i m_i\,,\label{E:equation2}\\
        w_i = T_i \alpha_i\,, T_i &= \prod_{j<i} (1 - \alpha_j)\,, \alpha_i = \mathcal{N} (\bm{u}; \bm{u}_i, \mathbf{\Sigma}^{\prime}_i ) \, o_i\,,\label{E:equation3}
\end{align}
where $\mathbf{M} \in \{\mathbf{I}, \mathbf{D}, \mathbf{N}, \mathbf{K} \}$ and $m_i \in \{\bm{c}_i, d_i, \bm{n}_i, k_i\}$ is the corresponding modality feature, with $d_i$ being the distance from the viewpoint centre to the intersection point of the camera ray and the Gaussian primitive $\bm{g}_i$; $w_i$ indicates the rendering contribution of $\bm{g}_i$ to pixel $\bm{u}$; and $\mathbf{\Sigma}^{\prime}_i$ and $\bm{u}_i$ are the primitive's covariance matrix and centre projected onto the image space~\citep{zwicker2001pv};
For more technical details about the rendering process, please refer to Gaussian surfel~\citep{dai2024siggraph}.

\subsection{Incremental Mapping} \label{SS:incremental_mapping}
We collect measurements captured at planned viewpoints and incrementally update our map representation. 
Given the current RGB image $\mathbf{I}^{\star}$ and depth image $\mathbf{D}^{\star}$ measurements, we generate a per-pixel point cloud using known camera parameters. 
We then update our voxel map $\mathcal{V}$ probabilistically based on the new point cloud, following OctopMap~\citep{hornung2013ar}. 

For GS map update, we first add Gaussian primitives to $\mathcal{G}$ where needed. 
To this end, we render the colour map $\mathbf{I}$, depth map $\mathbf{D}$, and opacity map $\mathbf{O}$ at the current camera viewpoint. 
We calculate a binary mask to identify the pixels that should be considered for densifying the GS map:
\begin{align}
    \mathbf{B}(\bm{u}) = &(\mathbf{O}(\bm{u})<0.5) \vee (\mathrm{avg}(\left| \mathbf{I}(\bm{u}) - \mathbf{I}^{\star}(\bm{u}) \right|) > 0.5)\notag\\
     &\vee (\mathbf{D}(\bm{u}) - \mathbf{D}^{\star}(\bm{u}) > \lambda \mathbf{D}^{\star}(\bm{u}))\,,\label{E:equation4}
\end{align}
where $\mathrm{avg}(\,\cdot\,)$ is the channel-wise averaging operation to calculate per-pixel colour error and $\lambda$ is a constant accounting for depth sensing noise, set to $0.05$ in our pipeline.
This mask indicates areas where opacity is low, colour rendering is inaccurate, or new geometry appears in front of the current depth estimate, signalling the need for new Gaussian primitives. 
We spawn new Gaussian primitives by unprojecting pixels on these areas into 3D space, with initial parameters defined by the corresponding point cloud position, pixel colour, and normal estimated by applying central differencing on the bilateral-filtered depth image~\citep{newcombe2011ismar}, which helps mitigate noise contained in the depth sensing. 
We also set scale values to $1$\,cm, opacity value to $0.5$, and confidence value to $0$.

At each mapping step, we train our GS map $\mathcal{G}$ using all collected RGB-D measurements for $10$ iterations. 
Specifically, for each iteration, we select the $3$ most recent frames and $5$ random frames from the measurement history. 
The loss for a frame $\{\hat{\mathbf{I}}\,,\hat{\mathbf{D}} \}$ in the training batch is formulated as the weighted sum of individual loss terms:
\begin{equation}
    \mathcal{L} = w_{c} \mathcal{L}_{c} + w_{d} \mathcal{L}_{d} + w_{n} \mathcal{L}_{n}\,,\label{E:equation5}
\end{equation} 
where the photometric loss $\mathcal{L}_c= L_1(\mathbf{I},\hat{\mathbf{I}})$ and the depth loss $\mathcal{L}_d = L_1(\mathbf{D},\hat{\mathbf{D}})$ are both calculated using the $L_1$ distance. The normal loss $\mathcal{L}_{n} = D_\mathit{cos}(\mathbf{N}, \widetilde{\mathbf{N}}) + TV(\mathbf{N})$ consists of the cosine distance $D_\mathit{cos}$ between the rendered normal map and the normal map $\widetilde{\mathbf{N}}$ derived from the rendered depth map~\citep{dai2024siggraph}, along with the total variation $TV$ loss~\citep{rudin1994icip} to enforce smooth normal rendering between neighbouring pixels. 
Note that the training process involves only a subset of the Gaussian primitive parameters $(\bm{x}_i, \bm{q}_i, \bm{s}_i, \bm{c}_i, o_i)$, while the modelling of non-trainable $k_i$ is introduced in~\cref{SS:confidence_modelling_for_gaussian_primitives}.

After every $5$ mapping steps, we perform a visibility check on all Gaussian primitives and delete those invisible to all history views to compact the GS map. 
We consider a Gaussian primitive visible from a viewpoint if at least one pixel rendered in that view receives its rendering contribution greater than a threshold, as defined in~\cref{E:equation3}.
Unlike previous works utilising density control during offline training~\citep{kerbl2023tog,dai2024siggraph,huang2024siggraph}, our approach adds necessary primitives and removes invisible ones during online missions, achieving computationally efficient scene reconstruction.

\subsection{Confidence Modelling for Gaussian Primitives} \label{SS:confidence_modelling_for_gaussian_primitives}
A Gaussian primitive can be effectively optimised 
if observed from different viewpoints. 
Based on this insight, we derive the confidence of a Gaussian primitive from the spatial distribution of viewpoints in the measurement history. 
Specifically, we connect the Gaussian centre $\bm{x}_i$ to the viewpoint centre $\bm{x}_{\bm{p}_j}$, denoted as $\bm{d}_{ij} = \bm{x}_{\bm{p}_j} - \bm{x}_i = d_{ij} \bm{v}_{ij}$, where $d_{ij}$ is the distance and $\bm{v}_{ij}$ is the normalised view direction, with $j \in \mathcal{S}(\bm{g}_i)$ and $\mathcal{S}$ being the index set of viewpoints from which the Gaussian primitive $\bm{g}_i$ has been observed. 
The confidence $k_i$ is finally formulated as:
\begin{align}
    k_i &= \gamma_i \, \mathrm{exp}(\beta_i)\,,\label{E:euqation6}\\
    \gamma_i &= \sum_{j \in \mathcal{S}(\bm{g}_i)} \left(1 - \frac{d_{ij}}{d_\text{far}}\right) \bm{n}_i \cdot \bm{v}_{ij}\,,\label{E:equation7}\\
    \beta_i &= 1 - \left\|\bm{\mu}_i\right\|\,,
    \bm{\mu}_i = \frac{1}{\left|\mathcal{S}(\bm{g}_i)\right|}\sum_{j \in \mathcal{S}(\bm{g}_i)} \bm{v}_{ij}\,,\label{E:equation8}
\end{align}
where $\gamma_i$ accounts for distance-weighted cosine similarity between the Gaussian primitive's normal $\bm{n}_i$ and view direction $\bm{v}_{ij}$, with $d_\text{far}$ as the maximum depth sensing range. Note that we increase the impact of viewpoints that are closer to the Gaussian primitive's centre or provide view directions similar to the primitive's normal.
$\beta_i$ measures the dispersion of directions from which $\bm{g}_i$ is observed, with $\beta_i$ closer to $0$ indicating similar view directions. 
Our confidence formulation assigns higher confidence to Gaussian primitives densely observed from viewpoints with varying angles,  whereas lower confidence to those with sparse and similar observations.

\subsection{Viewpoint Utility Formulation} \label{SS:viewpoint_utility_formulation}
Active scene reconstruction requires both exploration, to cover unknown areas, and exploitation, to closely inspect under-reconstructed surfaces.
In this work, we combine utility derived from the voxel map for exploration and the GS map for exploitation, enabling these behaviours effectively. 

A candidate viewpoint $\bm{p}_i^{c} \in \mathbb{R}^5$ is defined by its 3D position, yaw, and pitch angles in our framework. 
To simplify path planning, we constrain the positions to a discrete lattice placed at the centres of all free voxels. 
We follow existing active scene exploration frameworks~\citep{isler2016icra, schmid2020ral,bircher2016icra,zhou2021ral,palazzolo2018drones} and define the exploration utility based on the number of unexplored voxels visible from a candidate viewpoint.
Without relying on time-consuming ray-casting operations, our framework leverages efficient rendering of our GS map to identify visible voxels. 
We achieve this by checking whether the projected depth of the in-view voxel centres in the camera coordinate is smaller than the corresponding depth value in the rendered depth from the GS map. 

Combining unexplored region information in the voxel map and confidence rendering from the GS map, we define the utility of a candidate viewpoint $\bm{p}_i^{c}$ as: 
\begin{align}
    U_\text{view}(\bm{p}_i^{c}) = \phi \, U_{\mathcal{V}}(\bm{p}_i^{c}) + U_{\mathcal{G}}(\bm{p}_i^{c})\,,\label{E:equation9}
\end{align}
where $U_{\mathcal{V}}(\bm{p}_i^{c}) = \frac{N_u(\bm{p}_i^{c})}{\left|\mathcal{V}\right|}$ is the exploration utility, defined by the ratio of the number of visible unexplored voxels $N_u(\bm{p}_i^{c})$ to the total number of voxels in the voxel map; $U_{\mathcal{G}}(\bm{p}_i^{c}) = -\mathrm{mean}(\mathbf{K}_i)$ is the exploitation utility, calculated as the negative mean of the confidence map $\mathbf{K}_i$ rendered at $\bm{p}_i^{c}$ following~\cref{E:equation2}; and $\phi$ is the exploration weight.

\subsection{Viewpoint Sampling and Evaluation} \label{SS:view_sampling_and_evaluation}

\begin{figure}[!t]
  \includegraphics[width=\columnwidth]{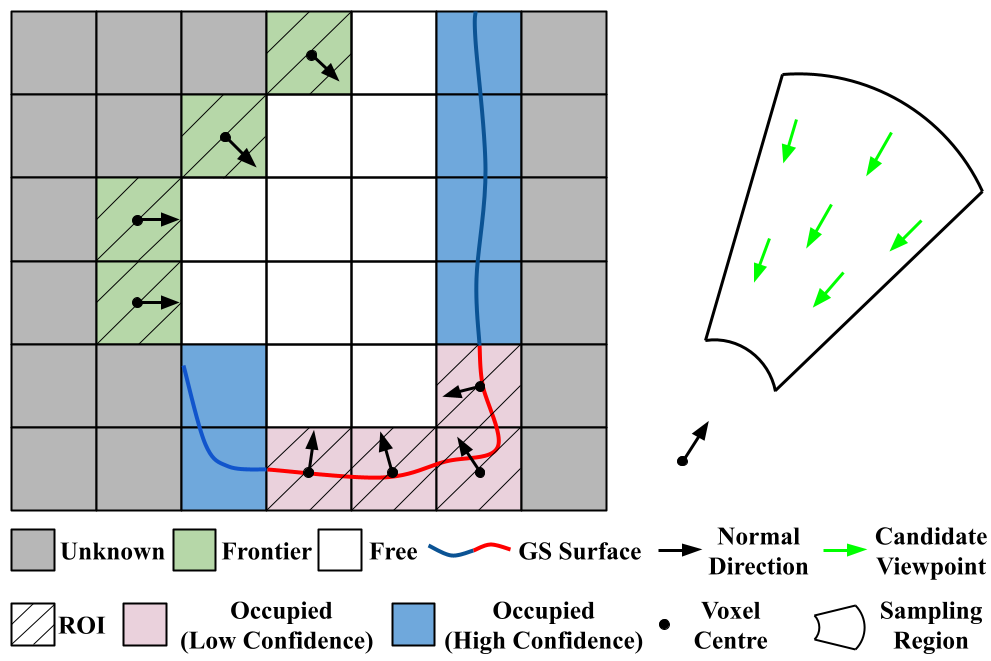}
  \caption{We show a 2D case of our ROI-based candidate viewpoint generation. 
  We identify ROI voxels by retrieving voxels containing low-confidence Gaussian primitives and frontier voxels. 
  Normals for low-confidence voxels are generated by averaging the normals of low-confidence primitives, while frontier voxel normals are calculated using average vectors to neighbouring free voxels.
  Given the voxel centres and directional normals, we generate candidate viewpoints within the sampling region, as illustrated on the right.
  }\label{F:sampling}
  \vspace{-0.1cm}
\end{figure}

Our viewpoint sampling strategy involves two types of candidate viewpoints. 
First, we randomly sample $N_\text{random}$ candidate viewpoints within a specified range around the current viewpoint.
However, relying solely on random local sampling can lead to local minima.
To address this, we introduce additional candidate viewpoints based on regions of interest (ROI) defined in the voxel map. 
We begin by identifying frontier voxels~\citep{yamauchi1997cira} and add them to our ROI set $\mathcal{R}$.
By explicitly modelling the confidence of each Gaussian primitive, we can identify and also include voxels containing low-confidence Gaussian primitives in $\mathcal{R}$. 
Inspired by previous work~\citep{kompis2021ral}, we define normals for each voxel in $\mathcal{R}$ to indicate the most informative viewing direction.
For voxels with low-confidence Gaussian primitives, this is simply the average normal of these Gaussian primitives. 
The normal of frontier voxels is determined by finding their neighbouring free voxels and calculating the average directional vector from the frontier voxel to these neighbours.
To generate ROI-based candidate viewpoints, we create a fixed number of candidate viewpoints within the cone defined by the minimum and maximum sampling distances to each ROI's centre, and the maximum angular difference relative to its normal.
Starting from the closest ROI voxel, we continue outward until we have collected up to $N_\text{ROI}$ viewpoints in free space.
We illustrate the sampling process in~\cref{F:sampling}.

We evaluate the utility of all candidate viewpoints following~\cref{E:equation9}. 
We use the $A^*$ algorithm~\citep{hart1968tssc} to find the shortest traversable path from the current viewpoint position to all candidate viewpoint positions.
Taking travel distance into account, we select the next best viewpoint $\bm{p^{\star}}$ by:
\begin{equation}
\bm{p}^{\star} = \underset{\bm{p}_i^{c}}{\mathrm{arg\,max}}\left(\frac{U_\text{view}(\bm{p}_i^{c})}{\sum_{i=1}^{N_\text{total}}U_\text{view}(\bm{p}_i^{c})} - \delta\frac{ U_\text{path}(\bm{p}_i^{c})}{\sum_{i=1}^{N_\text{total}}U_\text{path}(\bm{p}_i^{c})}\right),\label{E:equation10}
\end{equation}
where $N_\text{total} = N_\text{random} + N_\text{ROI}$; $U_\text{path}$ is the travel distance to the candidate viewpoint positions; and $\delta$ is a weighting factor for the travel cost.

\section{Experimental Evaluation} \label{S:experimental_evaluation}
Our experimental results support our three claims: (i) we show that our ActiveGS framework outperforms state-of-the-art NeRF-based and GS-based active scene reconstruction methods;
(ii) we show that our confidence modelling of Gaussian primitives enables informative viewpoint evaluation and targeted candidate viewpoint generation, improving reconstruction performance; (iii) we validate our framework in different simulation scenes and in a real-world scenario to show its applicability. 

\subsection{Implementation Details} \label{SS:implementation_details}
\textbf{Mapping}. 
We use a voxel size of $20$\,cm $\times$ $20$\,cm $\times$ $20$\,cm for the voxel map. 
We set the loss weights in~\cref{E:equation5} as: $w_c = 1.0$, $w_d = 0.8$, and $w_n = 0.1$. For visibility checks, a minimum rendering contribution threshold of $0.3$ is applied.

\textbf{Planning}.
We set the exploration weight $\phi = 1000$ in~\cref{E:equation9} to encourage exploratory behaviour during the initial phase of an online mission. 
We set $\delta =0.5$ for weighting travel costs in~\cref{E:equation10}. 
We consider $N_\text{total} = 100$ candidate viewpoints, including up to $N_\text{ROI} = 30$ ROI-based samples and $N_\text{random} = N_\text{total} - N_\text{ROI}$ random samples generated within $0.5$\,m of the current viewpoint position.

We test our implementation on a desktop PC with an Intel Core i9-10940X CPU and an NVIDIA RTX A5000 GPU.
In this setup, one mapping and planning steps take approximately $1$\,s and $0.5$\,s, respectively. 
The whole framework consumes \mbox{$4 - 5$}\,GB GPU RAM during an online mission, with approximately $10$\% allocated to the voxel map update.

\subsection{Simulation Experiments} \label{SS:simulation_experiments}
\begin{figure*}[!t]
    \centering
    \includegraphics[width=1.0\textwidth]{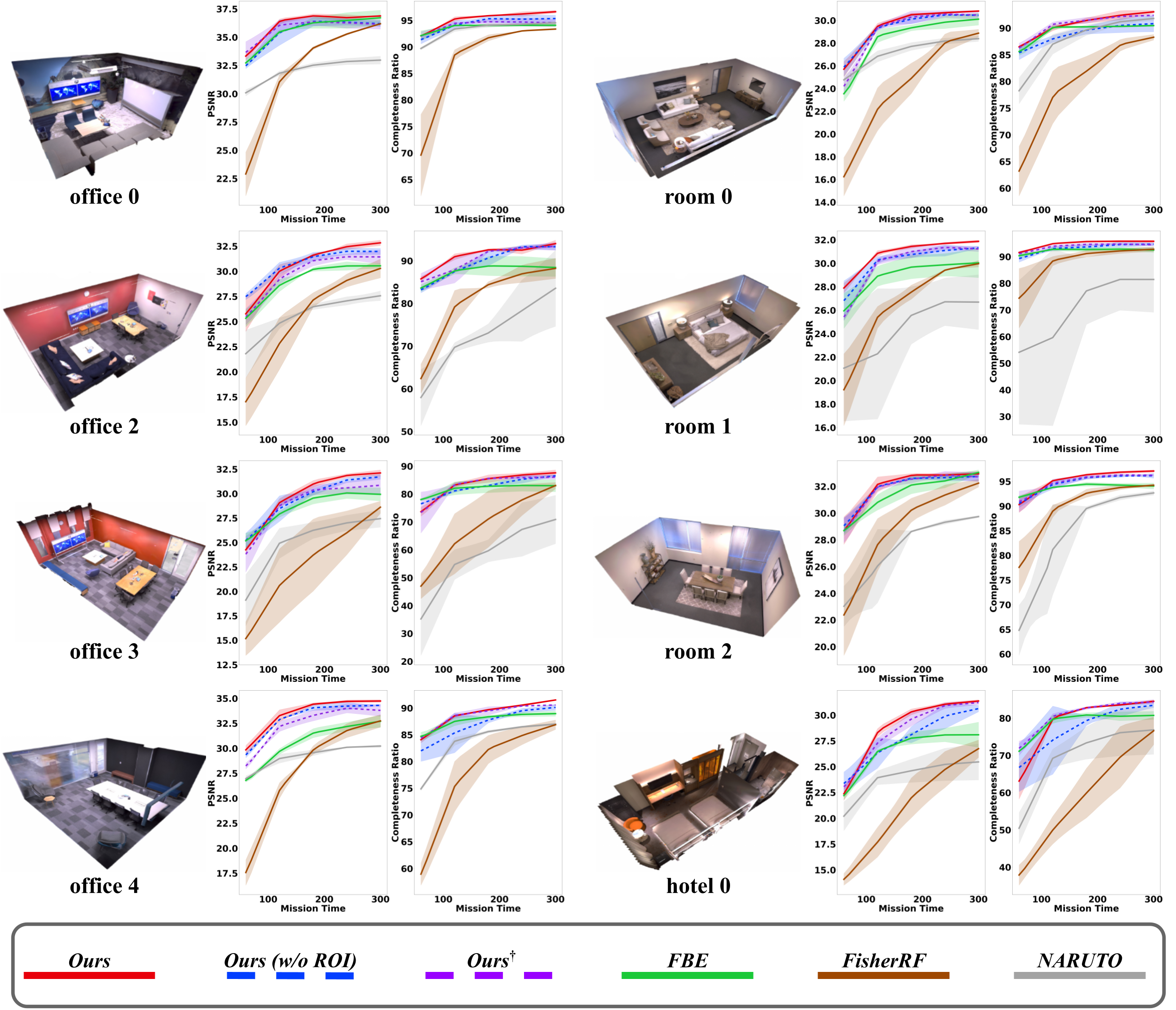}
    \caption{We report the reconstruction performance evaluated in rendering and mesh quality over online mission time. 
    Our ActiveGS outperforms baselines in all test scenes. 
    Our view planner considers unexplored regions for exploration, while exploiting low-confidence Gaussian primitives for further inspection. 
    Compared to GS-based approaches, our approach proposes explicit confidence modelling of Gaussian primitives, enabling targeted candidate viewpoint generation and fast viewpoint evaluation. 
    Our approaches demonstrate a large performance gain to the state-of-the-art NeRF-based approach, \textit{NARUTO}, motivating the use of GS in active scene reconstruction.} \label{F:quantative_results}
    \vspace{-0.2cm}
\end{figure*}
\textbf{Setup}.
We conduct our simulation experiments using the Habitat simulator~\citep{savva2019iccv} and the Replica dataset~\citep{straub2019arxiv}. 
The experiments utilise an RGB-D camera with a field of view of $[60^{\circ}, 60^{\circ}]$ and a resolution of $[512, 512]$ pixels.
The camera has a depth sensing range of $[0.1, 5.0]$\,m and Gaussian noise in the depth measurements with linearly increased standard deviation $\sigma = 0.01 d$, where $d$ represents the depth value.

\textbf{Evaluation Metrics}.
We report the reconstruction performance over total mission time, defined as the summation of mapping time, planning time, and action time, assuming a constant robot velocity of $1$\,m/s.
The reconstruction performance is evaluated on both rendering and mesh quality. 
For the rendering evaluation, we generate ground truth RGB images captured from $1000$ uniformly distributed test viewpoints in the scene's free space. 
We report PSNR~\citep{mildenhall2020eccv} of RGB images rendered from our GS map at test viewpoints as the rendering quality metric. 
For the mesh evaluation, we run TSDF fusion~\citep{newcombe2011ismar} on depth images rendered at training viewpoints and extract the scene mesh using Marching Cubes~\citep{lorensen1998sg}. 
We use the completeness ratio~\citep{feng2024cvpr} as the mesh quality metric with a distance threshold set to $2$\,cm.

We consider the following methods:
\begin{itemize}
    \item \textit{Ours}: Our full ActiveGS framework utilising both exploration and exploitation utility measures. 
    We consider ROI-based sampling to achieve targeted candidate viewpoint generation as described in~\cref{SS:view_sampling_and_evaluation}.
    \item \textit{Ours (w/o ROI)}: A variant of our ActiveGS that leverages only local random sampling, with \mbox{$N_\text{ROI} = 0$}.
    \item \textit{Ours$^{\dag}$}: A variant of our ActiveGS with an alternative confidence formulation, assigning higher confidence to Gaussian primitives with more visible viewpoints, without considering their spatial distribution.
    \item \textit{FBE}~\citep{yamauchi1997cira}: Frontier-based exploration framework that solely focuses on covering unexplored regions, without accounting for the quality of the GS map. We use the collected RGB-D data to update the GS map, similar to our framework. 
    \item \textit{FisherRF}~\citep{jiang2023eccv}: GS-based active scene reconstruction using frontier voxels for ROI-based candidate viewpoint generation and Fisher information for viewpoint evaluation. 
    We replace its 3D GS map with our 2D GS.
    \item \textit{NARUTO}~\citep{feng2024cvpr}: A state-of-the-art NeRF-based active scene reconstruction pipeline.
\end{itemize}
We run $5$ trials for all methods across $8$ test scenes. We set the maximum mission time to $300$\,s and evaluate reconstruction performance every $60$\,s. 
We report the mean and standard deviation for PSNR and completeness ratio.

\begin{figure*}[!t]
    \centering
    \includegraphics[width=0.99\textwidth]{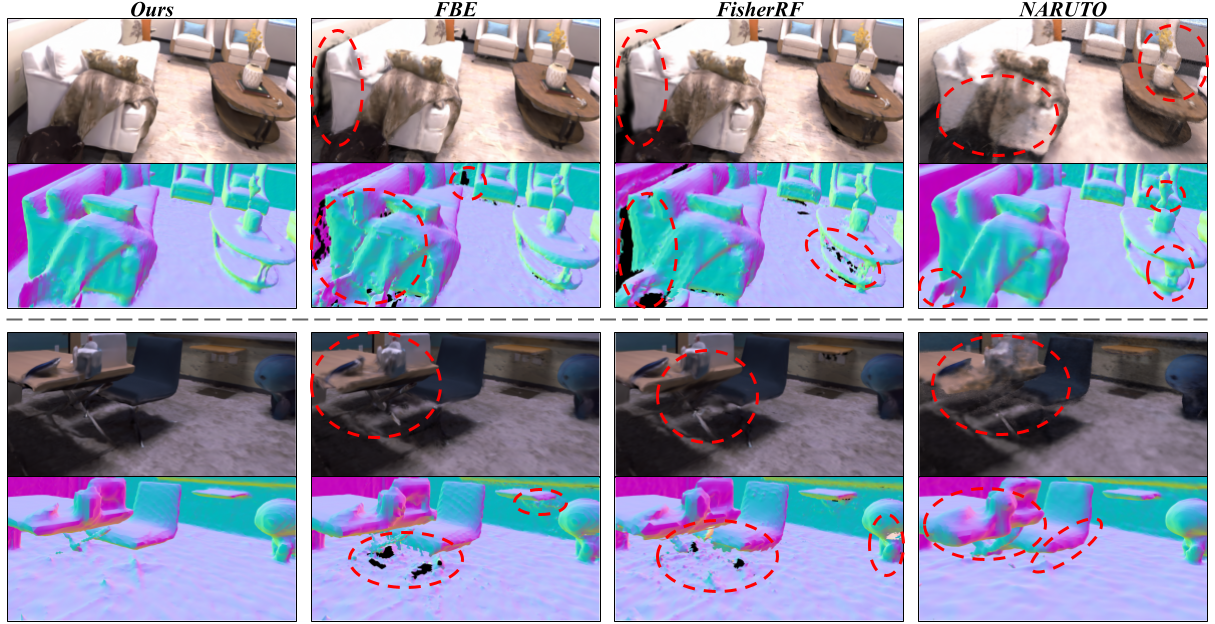}
    \caption{Visual comparison of reconstruction results using different approaches. We show RGB rendering and surface meshes for two scenes, with red circles highlighting areas of low-quality reconstruction from baseline approaches. Our ActiveGS considers both unexplored regions in voxel map and confidence value of GS map to enable targeted view planning, achieving complete and high-fidelity scene reconstruction.} \label{F:qualitative_results}
    \vspace{-0.3cm}
\end{figure*}

We present the results of simulation experiments in~\cref{F:quantative_results}. Our approach achieves the best performance in both rendering and mesh quality across all test scenes, supporting our first claim that it outperforms state-of-the-art NeRF and GS-based methods. 
The NeRF-based active scene reconstruction framework, \textit{NARUTO}, exhibits a significant performance gap compared to our approach, particularly in RGB rendering. 
This disparity arises because NeRF-based methods often compromise model capacity for faster map updates, limiting their representation quality in scene-level reconstruction. 
\textit{FisherRF} evaluates viewpoint utility by calculating the Fisher information in the parameters of the Gaussian primitives within its field of view. 
This requires computationally expensive gradient calculation for all candidate and training viewpoints, leading to prolonged planning times and incomplete reconstruction under limited mission time. 
Additionally, since Fisher information is conditioned on the candidate viewpoint, the viewpoint must be selected before its utility can be evaluated, preventing direct viewpoint sampling informed by Fisher information.
In contrast, our approach models the confidence of each Gaussian primitive, enabling fast feed-forward confidence rendering for viewpoint evaluation and identification of low-confidence surfaces for targeted candidate viewpoint generation, significantly enhancing reconstruction quality and efficiency.
\textit{FBE} focuses solely on exploration and ignores surface reconstruction quality, limiting its performance. 
While our approach balances exploration and exploitation by accounting for both unexplored regions and low-confidence Gaussian primitives. 
The ablation study comparing \textit{Ours} and \textit{Ours (w/o ROI)} demonstrates the benefits of ROI-based sampling for targeted inspection, reflected by higher means and smaller standard deviations in both evaluation metrics.
Our confidence formulation also outperforms the variant in \textit{Ours$^{\dag}$} by considering viewpoint distribution. 
These results confirm that our confidence modelling is effective in achieving efficient and high-fidelity active scene reconstruction, validating our second claim. 
We visually compare the reconstruction results in~\cref{F:qualitative_results}.

\subsection{Real-World Experiments} \label{SS:real-world_experiments}
We demonstrate the applicability our framework in a real-world experiment using a UAV equipped with an Intel RealSense 455 \mbox{RGB-D} camera to reconstruct a scene of size $6$\,m $\times$ $6$\,m $\times$ $3$\,m. 
Unlike simulation experiments, we do not account for the pitch angle of viewpoints in this experiment due to control limitations.
The UAV pose is tracked by an OptiTrack motion capture system. 
Given the limited on-board resources, we run ActiveGS on our desktop PC, where it receives \mbox{RGB-D} and pose data from the UAV for map updates and sends planned collision-free waypoints to guide the UAV. 
All communication is handled via ROS~\citep{quigley2009ros}.

Our real-world experiments indicate that our approach is effective for actively reconstructing unknown scenes by considering both unexplored regions in the voxel map and under-reconstructed surfaces in the GS map.
We show the experimental setup in~\cref{F:real-world_experiments} and the online active scene reconstruction in the supplementary video.
\begin{figure}[!t]
\centering
  \includegraphics[width=0.9\columnwidth]{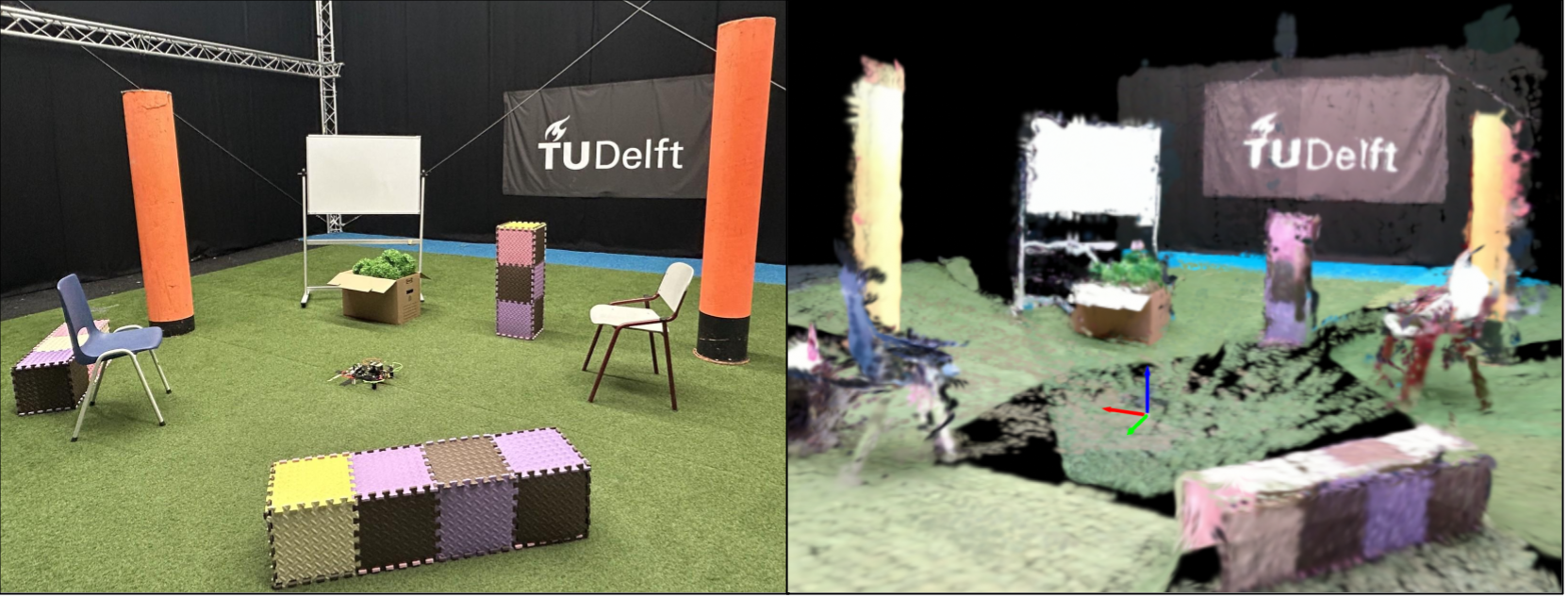}
  \caption{Our real-world experiments using a UAV equipped with an RGB-D camera. We show the experimental setup (left) and the RGB rendering from our GS map (right).} \label{F:real-world_experiments}
  \vspace{-0.2cm}
\end{figure}

\section{Conclusions and Future Work}\label{S:conclusions_and_future_work}
In this paper, we propose ActiveGS, a GS-based active scene reconstruction framework. 
Our approach employs a hybrid map representation, combining the high-fidelity scene reconstruction capabilities of Gaussian splatting with the spatial modelling strengths of the voxel map.
We propose an effective method for confidence modelling of Gaussian primitives, enabling targeted viewpoint generation and informative viewpoint evaluation. 
Our view planning strategy leverages the confidence information of Gaussian primitives to inspect under-reconstructed areas, while also considering unexplored regions in the voxel map for exploration.
Experimental results demonstrate that ActiveGS outperforms baseline approaches in both rendering and mesh quality.

A limitation of our current framework is the assumption of perfect localisation. Future work will incorporate localisation uncertainty into confidence modelling of Gaussian primitives. 
We also plan to better integrate the voxel map with GS map for more efficient mapping and leverage optimisation-based approaches to enhance view planning quality.

\section{Acknowledgment} \label{S:acknowlegment}
We thank Hang Yu, Jakub Plonka, and Moji Shi for their assistance in conducting the real-world experiments.

\bibliographystyle{IEEEtranSN}
\footnotesize
\bibliography{new}

\IfFileExists{./certificate/certificate.tex}{




\onecolumn

~\bigskip\bigskip\bigskip\bigskip 

	\section*{ \LARGE{Certificate of Reproducibility} }\vspace*{1cm}\Large{ 

	The authors of this publication declare that:}

	\vspace{5pt} 

	\begin{enumerate} 

		\setlength{\itemsep}{10pt} 

        \item The software related to this publication is distributed in the hope that it will be useful, support open research, and simplify the reproducability of the results but it comes without any warranty and without even the implied warranty of merchantability or fitness for a particular purpose.
  \item \textit{Liren Jin} primarily developed the implementation related to this paper. This was done on  Ubuntu20.04.

	\item \textit{Yue Pan} verified that the code can be executed on a machine that follows the software specification given in the Git repository available at: \\ 
\begin{center} 

  \url{https://github.com/dmar-bonn/active-gs} \end{center}

 \item \textit{Yue Pan} verified that the experimental results presented in this publication can be reproduced using the implementation used at submission, which is labeled with a tag in the Git repository and can be retrieved using the command:\\ 


\begin{center} 

  \verb|git checkout ral2025| 

\end{center} 

\end{enumerate} 

\twocolumn 

}{}

\end{document}